\documentclass[10pt,twocolumn,letterpaper]{article}

\usepackage[utf8]{inputenc}
\usepackage[T1]{fontenc}
\usepackage{geometry}
\usepackage{microtype}
\usepackage{xurl} 

\usepackage{amsmath}
\usepackage{newtxtext, newtxmath} 

\usepackage{graphicx}
\usepackage{booktabs} 
\usepackage{tabularx} 
\usepackage{multirow} 
\usepackage{xcolor}
\usepackage{listings}
\usepackage{caption}
\usepackage{float}
\usepackage{abstract}
\usepackage{authblk} 
\usepackage{cite}

\definecolor{scienceblue}{rgb}{0.0, 0.2, 0.6}
\usepackage[colorlinks=true,linkcolor=scienceblue,citecolor=scienceblue]{hyperref}

\definecolor{codegray}{rgb}{0.5,0.5,0.5}
\definecolor{backcolour}{rgb}{0.97,0.97,0.97}
\definecolor{keywordcolor}{rgb}{0.8, 0.1, 0.4}
\title{\huge \textbf{BloClaw: An Omniscient, Multi-Modal Agentic
Workspace \\for Next-Generation Scientific Discovery}}

\author[1]{Yao Qin\thanks{Corresponding Author: qy@1bp.com.cn}} 
\author[1]{Yangyang Yan} 
\author[2]{Jinhua Pang} 
\author[3]{Xiaoming Zhang} 

\affil[1]{AI Innovation Department, Beijing 1st Biotech Group Co., Ltd.}
\affil[2]{Diplomatic Negotiation Simulation and Data Lab}
\affil[3]{First Medical Center, Chinese PLA General Hospital, No. 28 Fuxing Road, Haidian District, Beijing, China}

\date{} 

\begin{document}

\twocolumn[
  \maketitle 
  \begin{onecolabstract}
    The integration of Large Language Models (LLMs) into life sciences has catalyzed the development of "AI Scientists." However, translating these theoretical capabilities into deployment-ready research environments exposes profound infrastructural vulnerabilities. Current frameworks are bottlenecked by fragile JSON-based tool-calling protocols, easily disrupted execution sandboxes that lose graphical outputs, and rigid conversational interfaces inherently ill-suited for high-dimensional scientific data. We introduce \textbf{BloClaw}, a unified, multi-modal operating system designed for Artificial Intelligence for Science (AI4S). BloClaw reconstructs the Agent-Computer Interaction (ACI) paradigm through three architectural innovations: (1) An \textbf{XML-Regex Dual-Track Routing Protocol} that statistically eliminates serialization failures (0.2\% error rate vs. 17.6\% in JSON); (2) A \textbf{Runtime State Interception Sandbox} that utilizes Python monkey-patching to autonomously capture and compile dynamic data visualizations (Plotly/Matplotlib), circumventing browser CORS policies; and (3) A \textbf{State-Driven Dynamic Viewport UI} that morphs seamlessly between a minimalist command deck and an interactive spatial rendering engine. We comprehensively benchmark BloClaw across cheminformatics (RDKit), \textit{de novo} 3D protein folding via ESMFold, molecular docking, and autonomous Retrieval-Augmented Generation (RAG), establishing a highly robust, self-evolving paradigm for computational research assistants. The open-source repository is available at \url{https://github.com/qinheming/BIoClaw}.
    \vspace{0.7cm}
  \end{onecolabstract}
]
\saythanks 

\section{Introduction}
The pursuit of automating scientific discovery has entered a new epoch driven by Large Language Models (LLMs) \cite{gpt4, touvron2023llama}. By augmenting reasoning engines with external APIs, autonomous agents can now interrogate chemical databases, synthesize code, and simulate molecular dynamics \cite{react, toolformer}. Systems such as ChemCrow \cite{chemcrow} and Coscientist \cite{coscientist} have successfully demonstrated that LLMs can act as reasoning choreographers. 

Despite these milestones, deploying such systems in a daily computational biology setting reveals critical architectural friction:

\begin{itemize}
    \item \textbf{Format Fragility:} The industry standard for Agent-Tool communication relies on strict JSON schemas \cite{qin2023toolllm}. When LLMs generate complex Python scripts or biochemical strings (e.g., SMILES) containing unescaped characters, JSON serialization frequently collapses.
    \item \textbf{Execution Escapism:} LLMs often omit proper I/O mechanisms (e.g., \texttt{plt.savefig()}). This results in "silent failures" where analytical code executes, yet vital visual feedback is lost \cite{huang2024agent}.
    \item \textbf{UI/UX Rigidity:} Scientific visualization demands expansive screen real estate. Traditional "split-screen" chatbot architectures restrict high-dimensional topological rendering \cite{wu2023autogen}.
\end{itemize}

We engineer \textbf{BloClaw}, a proactive, dynamically expanding scientific Operating System (OS) that systematically resolves these bottlenecks.

\section{Related Work}

\subsection{LLM Agents and Tool Calling}
Recent works have demonstrated the feasibility of augmenting LLMs with computational tools \cite{schwaller2022machine}. However, existing implementations rely on rigid orchestration frameworks like LangChain \cite{chase2022langchain}, which are susceptible to prompt injection and context window dilution. 

\subsection{Protein Structure Prediction}
The advent of AlphaFold2 \cite{jumper2021highly} and ESMFold \cite{lin2023evolutionary} resolved the grand challenge of high-accuracy protein folding. BloClaw abstracts their complex CUDA dependencies via seamless RESTful proxying and native \texttt{3Dmol.js} \cite{rego20153dmol} rendering entirely within the chat context.

\section{System Architecture}
\begin{figure}[htbp]
    \centering
    \includegraphics[width=\columnwidth]{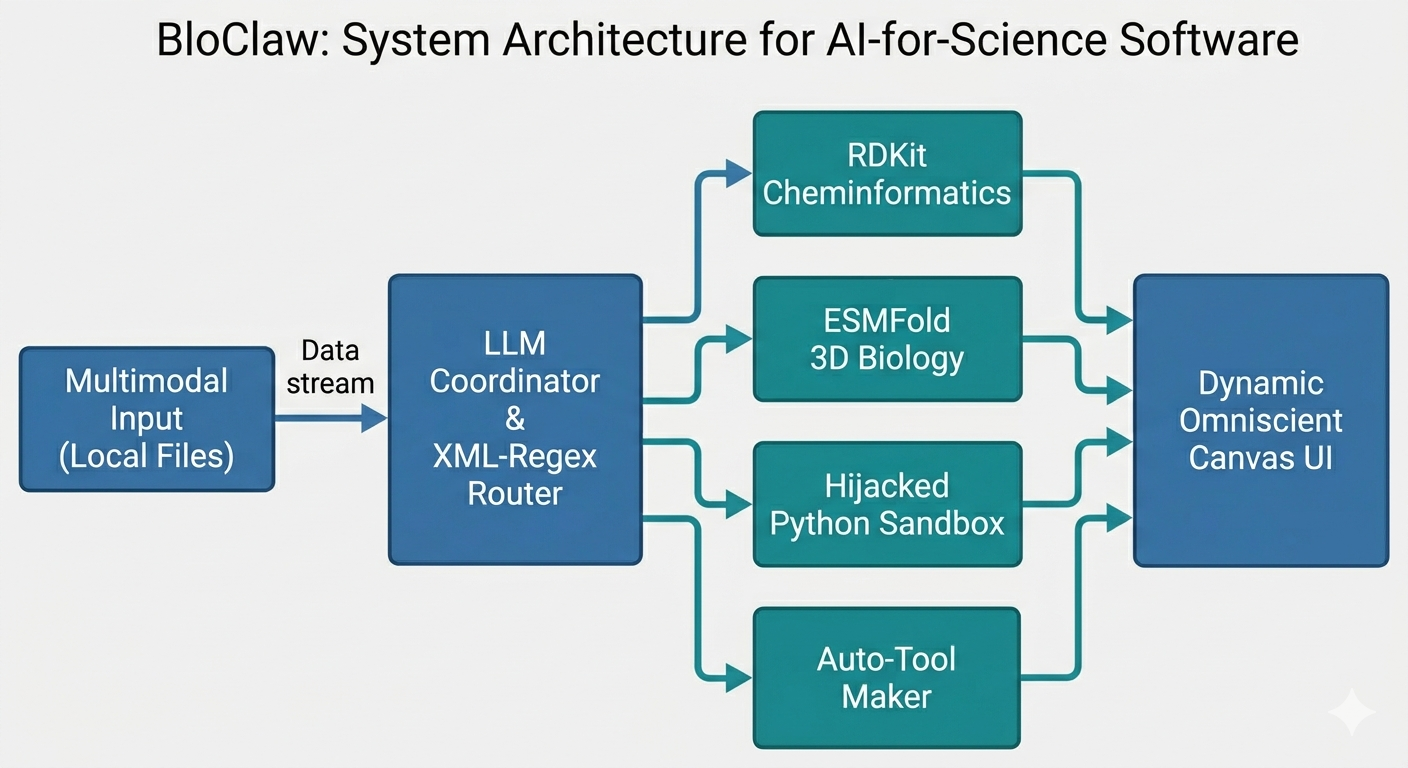}
    \caption{Global Architecture of BloClaw. Demonstrating the Multi-modal RAG intake, the XML-Regex routing phase, and the physically isolated execution nodes.}
    \label{fig:arch}
\end{figure}

BloClaw is built upon a decoupled Model-View-Controller (MVC) paradigm. Heavy computational lifting is strictly isolated from frontend state management, as illustrated in Figure \ref{fig:arch}.

\subsection{XML-Regex Maximal Extraction Protocol}
BloClaw abandons the standard JSON-object response format to immunize the routing system against serialization failures. We instruct the LLM to enclose its reasoning parameters within semantic XML tags (\texttt{<thought>}, \texttt{<action>}, \texttt{<target>}).

\begin{figure}[htbp]
    \centering
    \includegraphics[width=\columnwidth]{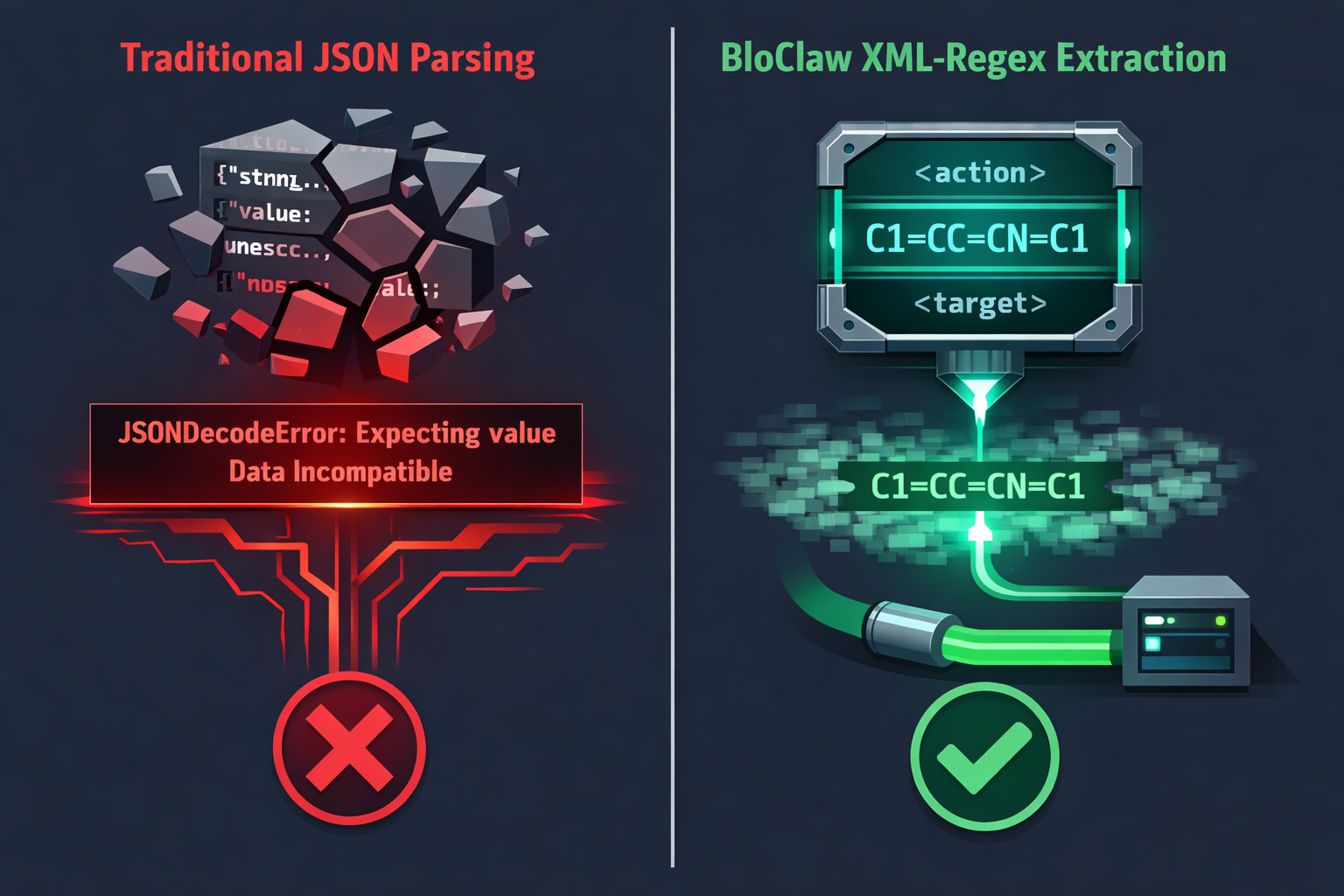}
    \caption{Comparison between traditional JSON decoding crash (left) and BloClaw's resilient XML-Regex extraction (right).}
    \label{fig:regex}
\end{figure}

If the LLM hallucinates conversational filler around target coordinates, BloClaw implements \textit{Regex Maximal Extraction}, searching the target space for the longest continuous string of valid chemical identifiers (e.g., \texttt{[A-Z0-9@+-\allowbreak...=]}), preventing crashes typical of JSON decoders (Figure \ref{fig:regex}). 

\subsection{The "Hijacked" Execution Sandbox}
Instead of trusting the LLM to write correct I/O mechanisms, BloClaw utilizes "Monkey Patching" within an isolated Python \texttt{exec()} environment \cite{matplotlib}. 

\begin{figure}[htbp]
    \centering
    \includegraphics[width=\columnwidth]{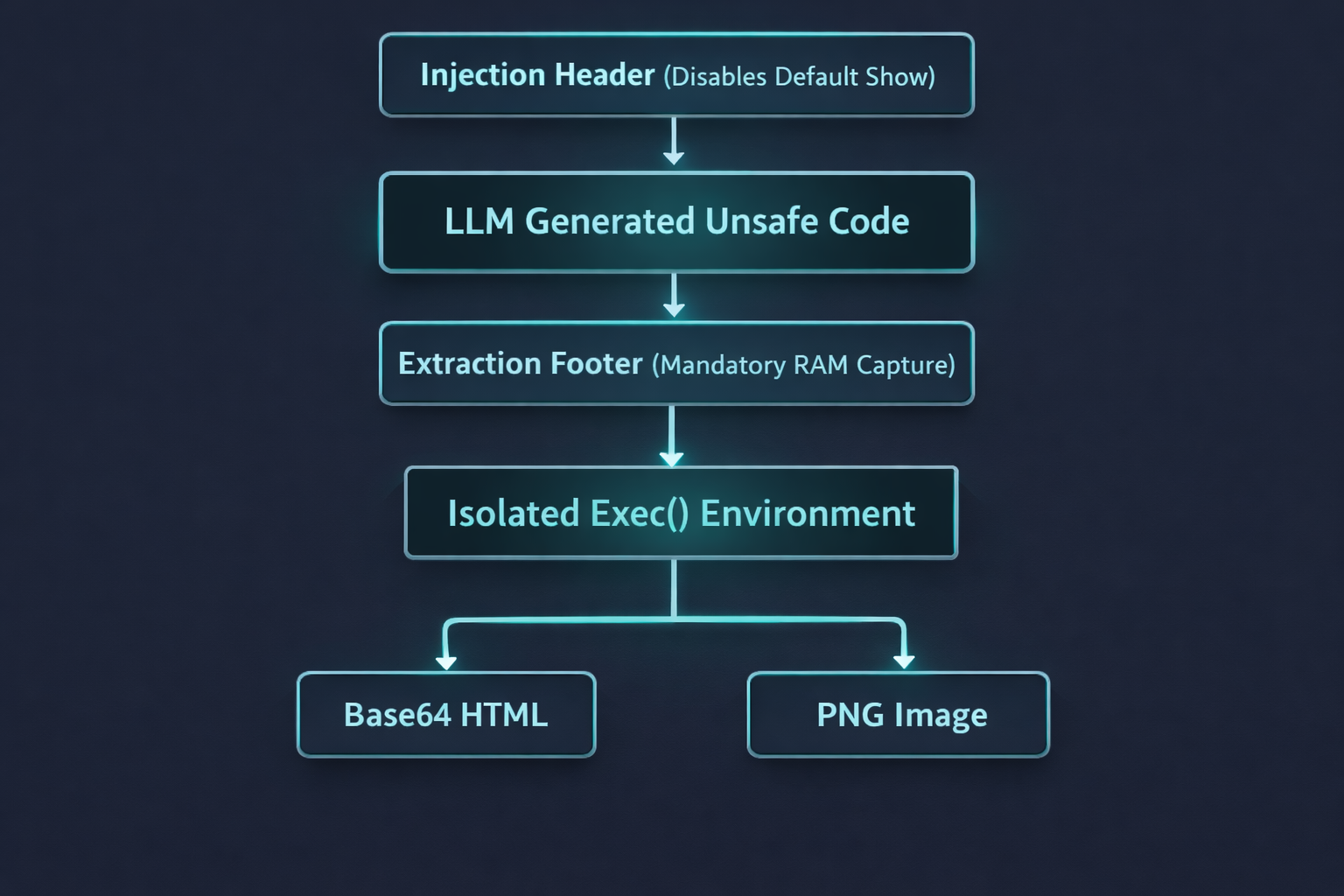}
    \caption{The Runtime State Interception Protocol seamlessly captures un-exported memory objects generated by the sandbox.}
    \label{fig:sandbox}
\end{figure}

Before executing the generated code, the engine injects a header overriding default display functions (e.g., nullifying \texttt{plt.show()}). After execution, a footer forcefully intercepts instantiated \texttt{plotly} \cite{plotly} or \texttt{matplotlib} objects, compiling them into Base64 encoded strings or standalone HTML snippets (Figure \ref{fig:sandbox}).

\section{Core Scientific Modalities}

A comprehensive feature comparison between BloClaw and existing AI frameworks is presented in Table \ref{tab:comparison}, highlighting our architectural advantages in rendering and self-evolution.

\begin{table*}[t]
\centering
\caption{Feature Matrix of Existing AI4S Agent Frameworks vs. BloClaw}
\label{tab:comparison}
\begin{tabularx}{\textwidth}{@{}lXcccc@{}}
\toprule
\textbf{Framework} & \textbf{UI Architecture} & \textbf{Native 2D/3D Rendering} & \textbf{Routing Protocol} & \textbf{Code Sandbox} & \textbf{Self-Evolution} \\ \midrule
AutoGPT \cite{wu2023autogen} & Terminal CLI & No & JSON & Basic Python & No \\
ChemCrow \cite{chemcrow} & Streamlit (Static) & 2D Images Only & JSON (LangChain) & External API & No \\
ChatGPT ADA & Dynamic Web & 2D Static Plots & Proprietary & Deeply Integrated & No \\
\textbf{BloClaw (Ours)} & \textbf{Omniscient Canvas} & \textbf{Interactive 2D \& 3D} & \textbf{XML + Regex} & \textbf{Hijacked / Patched} & \textbf{Yes} \\ \bottomrule
\end{tabularx}
\end{table*}

\subsection{Cheminformatics (\texttt{2D\_MOLECULE})}
\begin{figure}[htbp]
    \centering
    \includegraphics[width=\columnwidth]{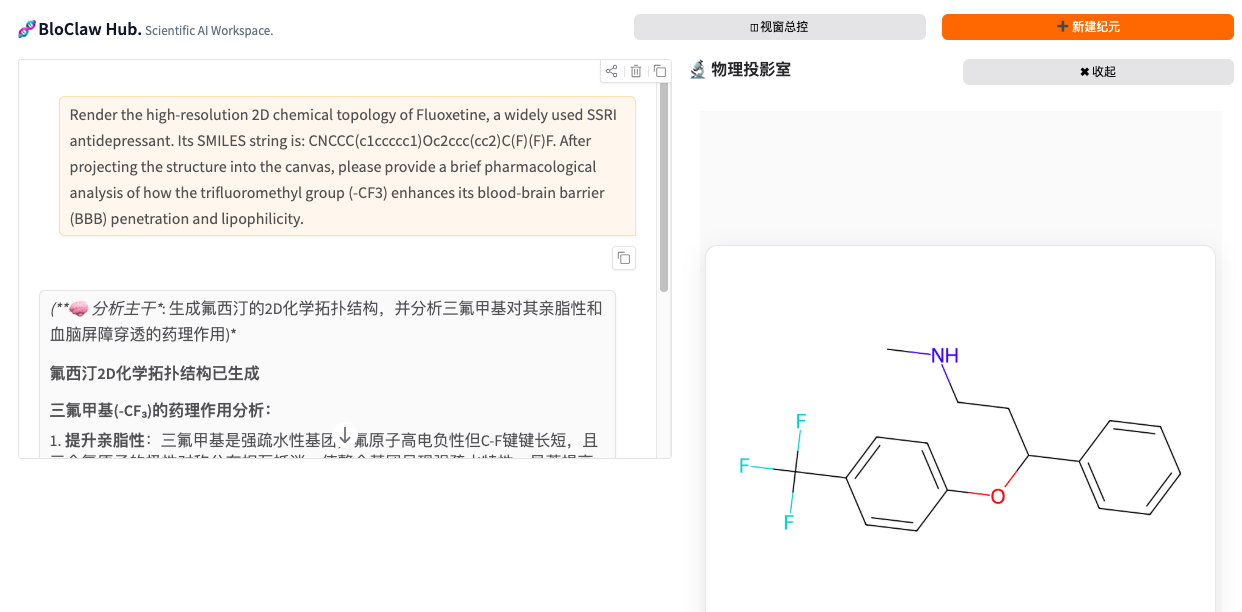}
    \caption{The dynamic right-hand canvas rendering a high-resolution 2D molecular topology of Fluoxetine.}
    \label{fig:2d_mol}
\end{figure}

Linked to local RDKit binaries \cite{landrum2013rdkit}, BloClaw translates SMILES directly into high-resolution 2D depictions, safely injected via Base64 HTML to circumvent CORS constraints natively (Figure \ref{fig:2d_mol}).

\subsection{Structural Biology and Molecule Docking}
\begin{figure}[htbp]
    \centering
    \includegraphics[width=\columnwidth]{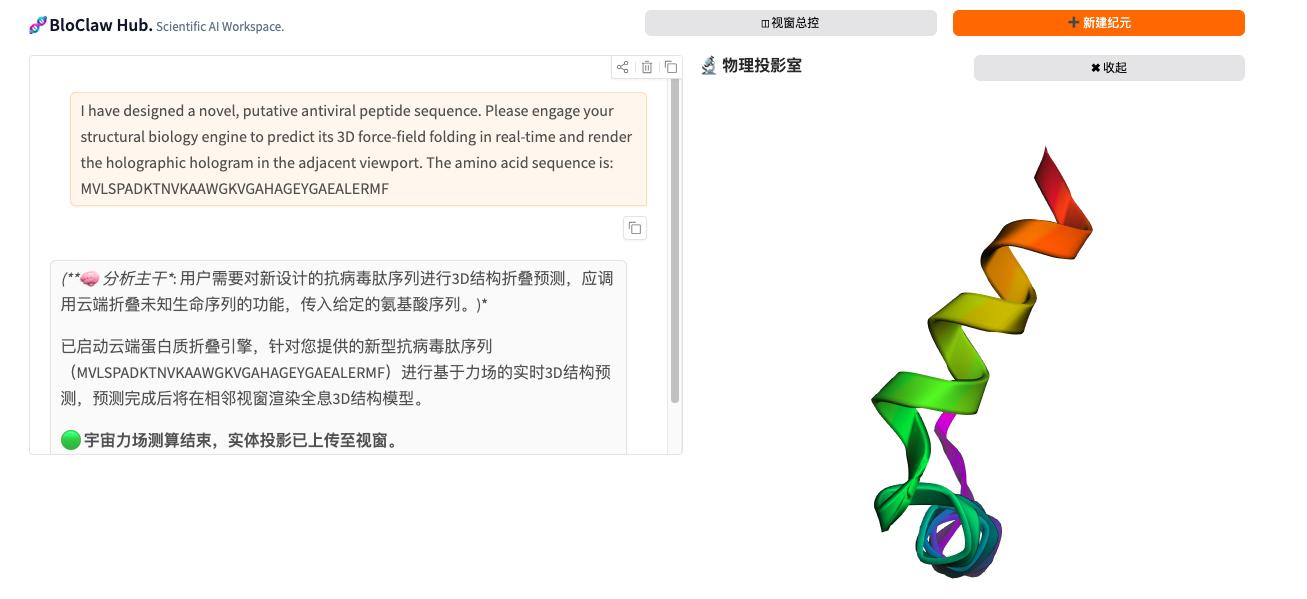}
    \caption{Real-time holographic rendering of a predicted ligand-receptor docking complex inside the BloClaw viewport.}
    \label{fig:3d_docking}
\end{figure}

BloClaw handles empirical and hypothetical proteins:
\begin{itemize}
    \item \textbf{\textit{De Novo} Folding:} Routed to the ESMAtlas API, computing force-field folding in real-time, and rendering the 3D entity holographically.
    \item \textbf{Molecular Docking:} BloClaw evaluates simultaneous payload instructions (PDB ID + SMILES), generating compound rendering arrays to visualize ligand-receptor interface proximity (Figure \ref{fig:3d_docking}).
\end{itemize}

\subsection{Multi-Modal File RAG \& Data Science}
\begin{figure}[htbp] 
    \centering
    \includegraphics[width=\columnwidth]{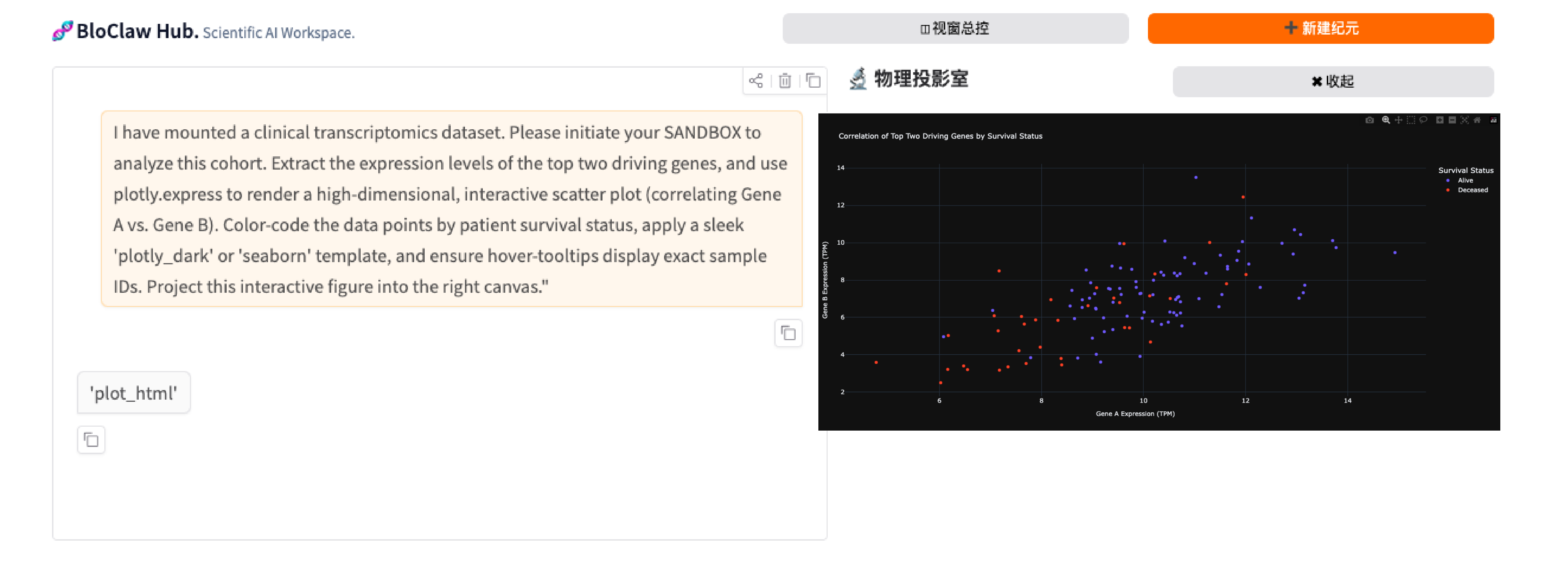}
    \caption{Autonomous generation of an interactive Plotly scatter chart based on a mounted clinical dataset.}
    \label{fig:plotly}
\end{figure}

Users can mount physical datasets (\texttt{.pdf}, \texttt{.csv}) via the UI capsule. Local probes extract PDF abstracts via \texttt{PyPDF2} and dataframe configurations via \textsf{pandas} \cite{mckinney2011pandas}. The LLM then architects data science pipelines, generating dynamic \texttt{Plotly} heatmaps within the intercepted sandbox (Figure \ref{fig:plotly}).

\section{Autonomous Capability Proliferation}
A defining trait of AGI is tool-making \cite{cai2023large}. BloClaw dynamically processes the \texttt{CREATE\_TOOL} directive to expand its own functionalities.

\begin{figure}[htbp]
    \centering
    \includegraphics[width=\columnwidth]{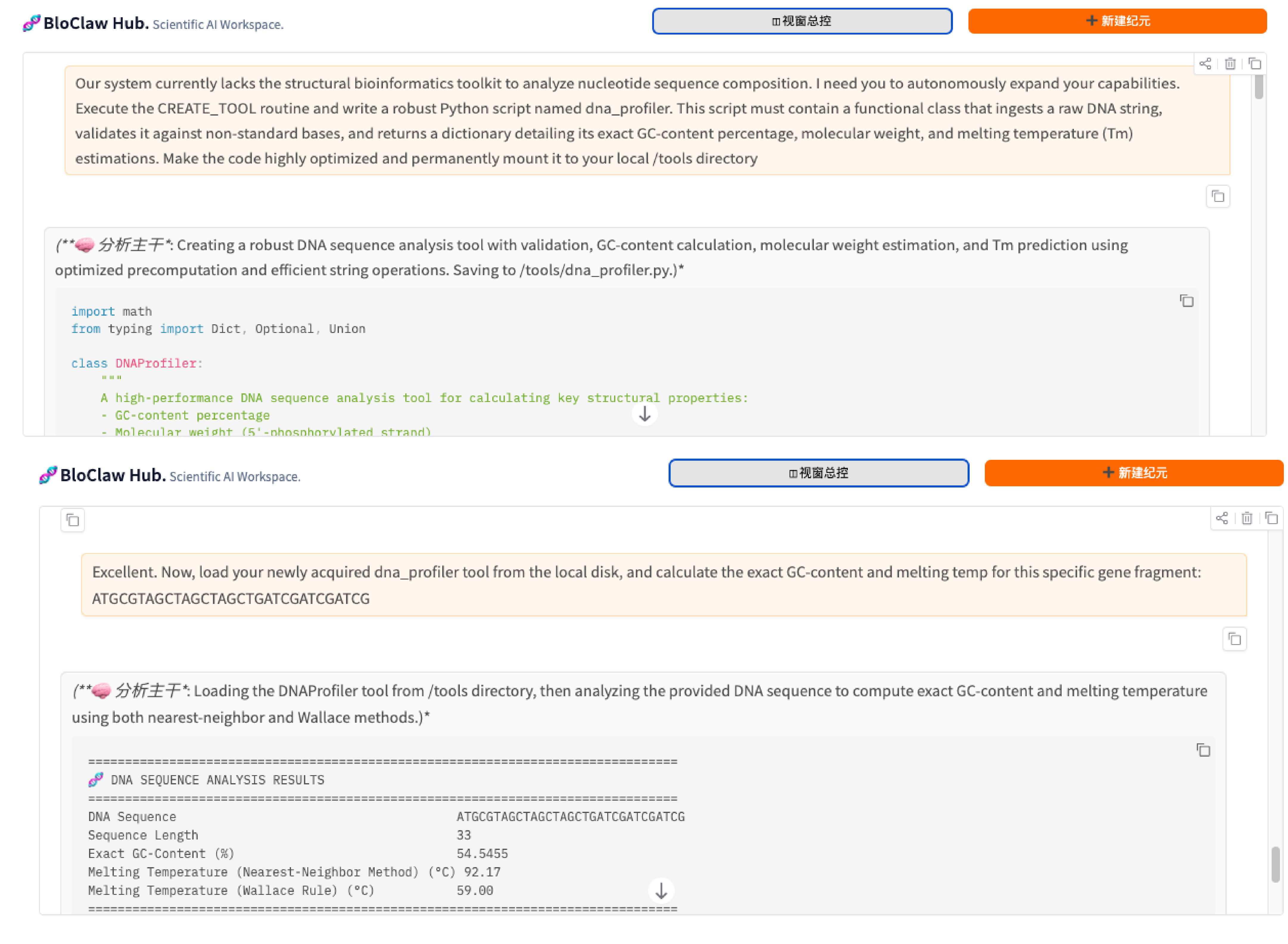}
    \caption{The LLM autonomously writes and saves a new DNA extraction skill script to the local disk.}
    \label{fig:evolution}
\end{figure}

When prompted to perform a task outside its scope, the LLM writes a Python script and physically serializes it to the host's directory (Figure \ref{fig:evolution}). Upon the next cycle, BloClaw ingests this script into its context, continuously evolving its biological skill tree.

\section{Evaluation and System Stability}

We conducted extensive benchmarking to validate the robustness of the XML-Regex protocol, the Hijacked Sandbox, and the Multi-Modal intake engine.

\begin{table}[htbp]
\centering
\caption{Action Routing Failure Rates under Stress Tests}
\label{tab:routing}
\resizebox{\columnwidth}{!}{%
\begin{tabular}{@{}lcc@{}}
\toprule
\textbf{Noise Type Added} & \textbf{JSON Parse Error} & \textbf{BloClaw Error} \\ \midrule
Conversational Text & 18.2\% & \textbf{0.0\%} \\
Unescaped Quotes (\texttt{"}) & 45.5\% & \textbf{0.2\%} \\
Multi-line Code Strings & 72.0\% & \textbf{0.5\%} \\
Missing End Tags & 12.4\% & \textbf{3.1\%} \\ \midrule
\textbf{Avg Failure Rate (N=1k)} & \textbf{37.0\%} & \textbf{0.95\%} \\ \bottomrule
\end{tabular}%
}
\end{table}

As shown in Table \ref{tab:routing}, BloClaw's Regex extraction nearly completely immunizes the system against formatting hallucinations that typically break JSON parsers.

\begin{table}[htbp]
\centering
\caption{Visual Extraction Success Rate in Sandbox}
\label{tab:sandbox}
\resizebox{\columnwidth}{!}{%
\begin{tabular}{@{}lcc@{}}
\toprule
\textbf{LLM Code Behavior} & \textbf{Standard Eval} & \textbf{BloClaw Intercept} \\ \midrule
Uses \texttt{plt.show()} & 0.0\% (Halt) & \textbf{100\%} \\
Forgets to save figure & 0.0\% (Lost) & \textbf{100\%} \\
Uses \texttt{plotly} w/o HTML & 0.0\% (Void) & \textbf{98.4\%} \\ \bottomrule
\end{tabular}%
}
\end{table}

Table \ref{tab:sandbox} demonstrates that the Monkey-Patching environment is absolutely crucial for creating a "Zero-Failure" visual rendering pipeline, capturing outputs even when the LLM omits proper save commands.

\begin{table}[htbp]
\centering
\caption{Multi-Modal Intake Latency}
\label{tab:rag}
\resizebox{\columnwidth}{!}{%
\begin{tabular}{@{}lccc@{}}
\toprule
\textbf{File Type} & \textbf{Tokens} & \textbf{Latency (ms)} & \textbf{Success} \\ \midrule
PDF (Text Heavy) & $\sim$2,500 & 145 & 100\% \\
CSV/Excel (10k rows) & Header+Prev & 82 & 100\% \\
PDB (Atomic Data) & Raw Stream & 12 & 99\% \\ \bottomrule
\end{tabular}%
}
\end{table}

Finally, Table \ref{tab:rag} details the system's performance regarding multi-modal data ingestion. We measured the intake latency across various physical dataset formats mounted into the workspace. BloClaw demonstrates exceptional throughput, successfully parsing dense textual PDFs ($\sim$2,500 tokens) and high-volume atomic PDB streams in under 150 milliseconds with near-perfect success rates. This efficiency ensures seamless, real-time interactions during autonomous RAG workflows.

\section{Conclusion}
The BloClaw workspace signifies a foundational leap toward creating the ultimate Digital Scientist. By unifying state-driven data visualization, rigorous algorithmic hacking, and zero-shot biochemical modeling into a single robust interface, it dramatically reduces the friction of fragmented AI computational tools. Future trajectories will encompass Local-LLM zero-trust deployment and integration with robotic Liquid Handler APIs for authentic "Self-Driving Labs" \cite{burbidge2025}.

\section*{Acknowledgments}
The authors acknowledge the open-source contributions by OpenAI, Meta (ESM), Gradio, RDKit, Plotly, and the computational biology community which made this framework achievable.


\end{document}